\documentclass[sigconf]{acmart}

\AtBeginDocument{%
  \providecommand\BibTeX{{%
    \normalfont B\kern-0.5em{\scshape i\kern-0.25em b}\kern-0.8em\TeX}}}

\copyrightyear{2025}
\acmYear{2025}
\setcopyright{acmlicensed}\acmConference[MM '25]{Proceedings of the 33rd
ACM International Conference on Multimedia}{October 27--31,
2025}{Dublin, Ireland}
\acmBooktitle{Proceedings of the 33rd ACM International Conference on
Multimedia (MM '25), October 27--31, 2025, Dublin, Ireland}
\acmDOI{10.1145/3664647.3681102}
\acmISBN{979-8-4007-0686-8/24/10}

\usepackage[framemethod=tikz]{mdframed}
\usepackage{amsmath}
\usepackage{amsfonts} 
\usepackage{latexsym}
\usepackage{graphicx}
\usepackage{stackengine}
\usepackage{booktabs}
\usepackage[utf8]{inputenc}
\usepackage[table]{colortbl,color}
\usepackage{xcolor}
\usepackage{multirow}
\usepackage{enumitem}
\usepackage{algorithm}
\usepackage{algpseudocode}

\usepackage{pifont}

\renewcommand\footnotetextcopyrightpermission[1]{}
\begin{document}

\title{Seeing Beyond the Scene: Enhancing Vision-Language Models with Interactional Reasoning}


\author{Dayong Liang}
\email{ft_ldy@mail.scut.edu.cn}
\affiliation{\country{South China University of Technology}}
\affiliation{\country{Peng Cheng Laboratory}}

\author{Changmeng Zheng}
\email{changmeng.zheng@polyu.edu.hk}
\affiliation{ \country{The Hong Kong Polytechnic University}}

\author{Zhiyuan Wen}
\email{wenzhy@pcl.ac.cn}
\affiliation{\country{Peng Cheng Laboratory}}

\author{Yi Cai}
\email{ycai@scut.edu.cn}
\affiliation{\country{South China University of Technology}}

\author{Xiao-Yong Wei}
\authornote{Corresponding Author}
\email{x1wei@polyu.edu.hk}
\affiliation{\country{The Hong Kong Polytechnic University}}

\author{Qing Li}
\email{qing-prof.li@polyu.edu.hk}
\affiliation{\country{The Hong Kong Polytechnic University}}


\begin{abstract}
Traditional scene graphs primarily focus on spatial relationships, limiting vision-language models' (VLMs) ability to reason about complex interactions in visual scenes. 
This paper addresses two key challenges: (1) conventional detection-to-construction methods produce unfocused, contextually irrelevant relationship sets, and (2) existing approaches fail to form persistent memories for generalizing interaction reasoning to new scenes. 
We propose Interaction-augmented Scene Graph Reasoning (ISGR), a framework that enhances VLMs' interactional reasoning through three complementary components. 
First, our dual-stream graph constructor combines SAM-powered spatial relation extraction with interaction-aware captioning to generate functionally salient scene graphs with spatial grounding. 
Second, we employ targeted interaction queries to activate VLMs' latent knowledge of object functionalities, converting passive recognition into active reasoning about how objects work together. 
Finally, we introduce a lone-term memory reinforcement learning strategy with a specialized interaction-focused reward function that transforms transient patterns into long-term reasoning heuristics. 
Extensive experiments demonstrate that our approach significantly outperforms baseline methods on interaction-heavy reasoning benchmarks, with particularly strong improvements on complex scene understanding tasks. 
The source code can be accessed at \textcolor{magenta}{\url{https://github.com/open_upon_acceptance}.}
\end{abstract}

\ccsdesc[500]{Computing methodologies~Scene Understanding}
%
\keywords{scene understanding; interactional reasoning; vision language models}

\maketitle

\section{Introduction}

\begin{figure*}
    \centering
    \includegraphics[width=6.3in]{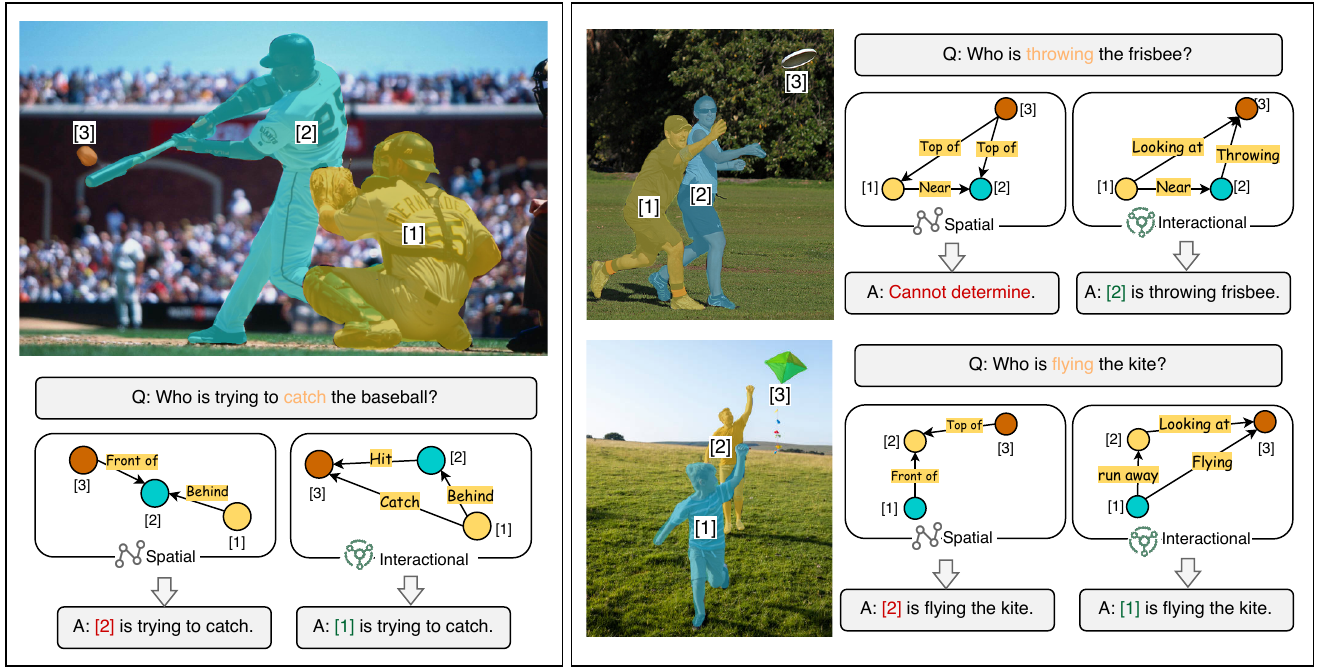}
    \caption{Examples showing how our interaction-augmented scene graphs enhance reasoning on dynamic interactions. Spatial: a conventional spatial-only scene graph misinterprets the situation as merely ``baseball in front of player''. Interaction: our approach correctly identifies the functional relationship ``player catch baseball'', enabling more accurate answer to the query ``Who is trying to catch the baseball?''.}
    \vspace{-0.1cm}
    \label{fig:introduction}
\end{figure*}

Scene graphs have been widely used to support multimodal reasoning tasks, such as image captioning, visual grounding, and visual question answering (VQA)\cite{chang2021comprehensive}. 
However, current scene graph construction methods primarily focus on positional or spatial relationships (e.g., ``on'', ``under'', ``next to''). 
This focus stems from the ease of annotating such relationships and the availability of well-established detection-to-construction frameworks for extracting them. 
As illustrated in Figure~\ref{fig:introduction}, while these spatial relationships are helpful for object-centric queries, they fall short in addressing more general user queries that often involve interactional or functional relationships (e.g., ``looking at'', ``Catching'', ``Throwing''). 
Such relationships are particularly important for causal reasoning, where distinguishing subjects and objects from distractors is critical.

This limitation significantly impacts the reasoning capabilities of vision-language models (VLMs), as most existing methods use constructed scene graphs as external sources for in-context learning without explicitly modeling interactions \cite{zheng2024picture,mitra2024compositional,mondal2024kam}. 
Recently, a new paradigm of scene graph-based reasoning has emerged, prompting VLMs to infer scene regions and use the results as more nuanced evidence for reasoning\cite{gu2024conceptgraphs, huang2024structure}. 
While this represents a step forward in incorporating interactional reasoning, the evidence remains coarse and fails to distinguish between subjects and objects within interactions. 
Additionally, these methods struggle to focus on contextually relevant concepts in the presence of distractors.
Therefore, existing approaches are limited in enabling models to form long-term memories, which are essential for generalizing interactional reasoning to new or unseen data. 
This limitation highlights the need for more advanced methods to capture and utilize subtle, interaction-based scene evidence effectively.

In this paper, we aim to enhance the interactional reasoning capabilities of vision-language models (VLMs) by enabling the construction of subtle interaction-augmented scene graphs and incorporating long-term memory reinforcement as

\noindent\textbf{Summarize-and-Align Graph Construction}: 
Constructing such graphs poses significant challenges, as inferring subtle relationships requires understanding complex contextual features such as intent, motion, or temporal dynamics. 
Unlike spatial relationships, which can often be derived from object positioning, many interactional relationships lack explicit visual cues, making them harder to detect.
To address this, we move away from the conventional detection-to-construct paradigm and introduce a summarize-and-align approach for graph construction. 
The key idea is to reduce focus drift by guiding VLMs to generate a disambiguated, contextually relevant, and focus-enhanced summarization of the image content. 
This summarization serves as a blueprint for generating an initial scene graph, which primarily captures easily detectable spatial relationships. 
By confining the scope of graph construction to contextually relevant concepts derived from the summarization, we avoid the diverse and often noisy outputs associated with open-ended conventional methods.
Furthermore, instead of relying on human annotations for subtle relationships, we propose an interactional chain-of-thoughts (ICoT) approach. 
This method encourages VLMs to reason over the initial graph by iteratively identifying the subjects and objects of each interaction. 
It aligns spatial relationships with interactional and functional relationships, enabling a richer and more nuanced understanding of the scene. 
This approach not only improves the granularity of interactional reasoning but also lays the foundation for generalizing to unseen data.

\noindent\textbf{Long-term memory reinforcement}: One of the key challenges in enabling long-term memory formation for VLMs in existing methods is the lack of annotated datasets containing subtle interactional relationships to use as tuning pairs. 
The proposed interaction-augmented graph helps bridge this gap, and fine-tuning can be easily performed using supervised fine-tuning (SFT). 
However, because the relationships in the graph are primarily generated through the ICoT rather than explicit human annotations, simple SFT alone is insufficient to guarantee high-quality memory formation.
To address this, we customize the Group Policy Optimization (GRPO) framework by introducing a reward mechanism. 
Rewards are assigned to successful interactional reasoning steps within the ICoT process, providing feedback that reinforces VLMs' ability to infer interactional relationships accurately.
By incorporating this reward-driven reinforcement, the graph construction process and the VLM inference module are unified within the same optimization loop. 
This collaborative approach ensures improved performance across both steps while also facilitating the formation of high-quality long-term memory for reasoning over subtle interactions.

\begin{figure*}
    \centering
    \includegraphics[width=6.8in]{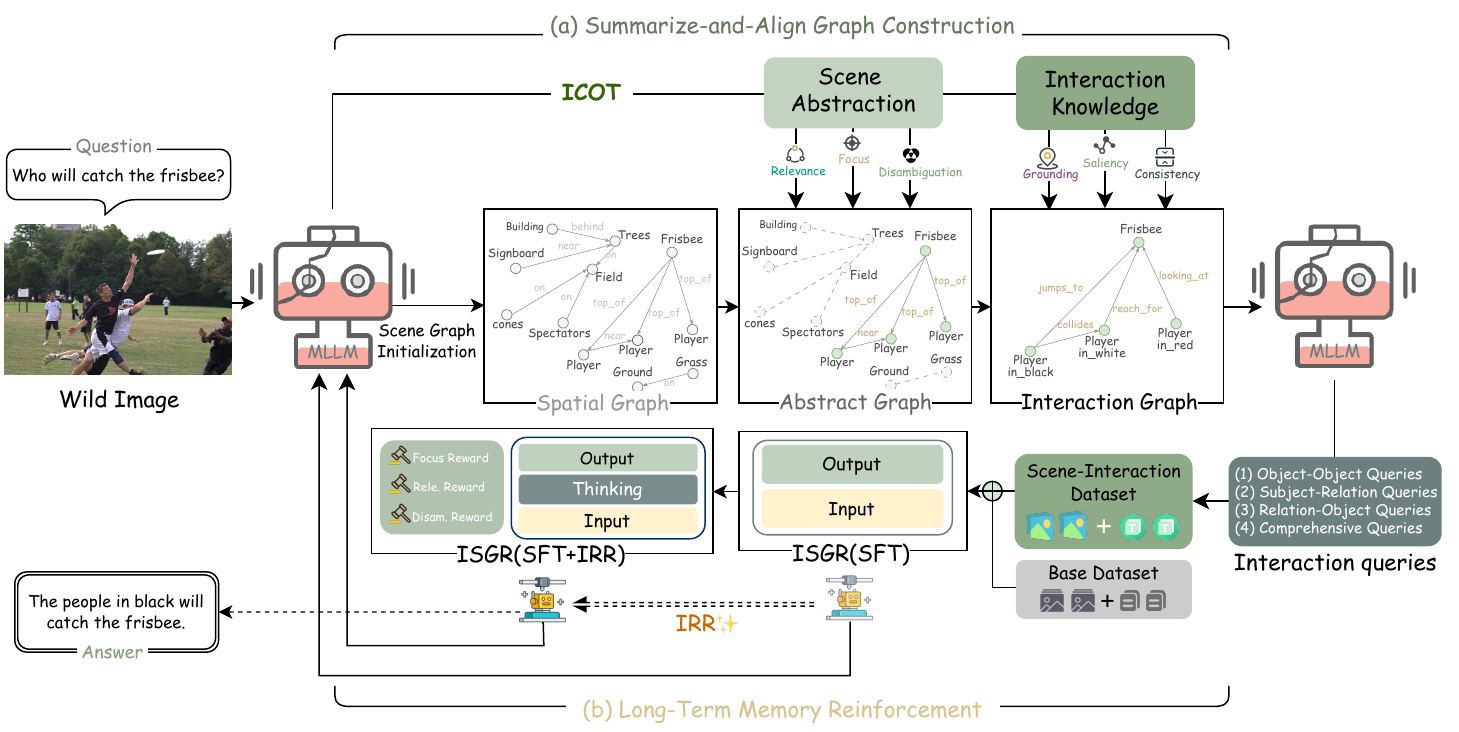}
    \caption{Overview of our Interaction-augmented Scene Graph Reasoning (ISGR) framework: (a) Summarize-and-Align Graph Construction transforms input images through Scene Graph Initialization and ICOT, progressively creating Spatial, Abstract, and Interaction Graphs with relevance, focus, and disambiguation constraints; (b) Long-Term Memory Reinforcement combines ISGR(SFT) and ISGR(SFT+IRR) models with Interaction Reasoning Reinforcement to enhance interaction reasoning capabilities on complex visual questions.}
    \vspace{-0.1cm}
    \label{model}
\end{figure*}

\section{Related Works}

\subsection{Instruction Tuning} 
A key challenge for large language models (LLMs) is the misalignment between their training objective—minimizing word prediction error—and users' expectation for helpful instruction adherence \cite{fedus2022switch,rae2021scaling,thoppilan2022lamda}. Instruction tuning effectively bridges this gap by training on (INSTRUCTION, OUTPUT) pairs, which shifts models beyond simple next-word prediction \cite{krishna2017visual,wang2023self,peng2023instruction,zhang2023video}. These datasets typically incorporate annotated natural language data, providing explicit task guidance \cite{longpre2023flan,xu2024vision}, or LLM-generated outputs from curated instructions, enhancing the quality of interactions \cite{zheng2023judging,chen2024allava,chen2024sharegpt4v}. However, instruction tuning primarily refines communication rather than imparts new knowledge, as studies suggest that LLMs acquire most of their capabilities during pretraining \cite{zhou2023lima,jiang2024survey}. Our work emphasizes the importance of aligning instruction data with human cognitive patterns \cite{lu2023instag, liu2023makes}, while maintaining structured information, which enables models to better understand scene interactions through human-like reasoning, instead of merely memorizing factual knowledge. By focusing on this alignment, we aim to improve the models' utility in real-world applications, ensuring they respond more effectively to user queries.

\vspace{-10pt}
\subsection{Scene Graph Generation} 
Scene graphs offer an ideal scaffold for structured interaction reasoning by capturing spatial and semantic relationships within visual environments \cite{chen2024scene,zhou2023vlprompt}. Since its introduction \cite{krishna2017visual} for image retrieval \cite{johnson2015image,schuster2015generating}, Scene Graph Generation (SGG) has evolved into a core component of structured visual understanding, with various approaches developed to address its challenges. Two-stage pipelines \cite{zheng2023prototype,lin2020gps,lin2022ru} separate object detection and relation classification, while one-stage methods \cite{cong2023reltr,liu2021fully,li2022sgtr} directly generate scene graphs. Additionally, open-vocabulary SGG \cite{he2022towards,yu2023visually,zhang2023learning} enables predicate recognition over unseen object categories by leveraging vision-language alignment. Despite these advancements, most existing SGG models are not designed for downstream instruction generation. Our approach uniquely utilizes fine-grained and grounded scene graphs as an intermediate representation to generate structured instruction-response data, thereby enhancing VLMs' ability to reason about interactions.

\subsection{SG-augmented VLMs} 
Recent approaches have explored integrating scene graphs into vision-language models to enhance relational reasoning. Parameter-heavy methods like MR-MKG \cite{lee2024multimodal}, Structure-CLIP \cite{huang2024structure}, and LLAVA-SG \cite{wang2025llava} incorporate additional modules to process graph structures, but often introduce complexity and may disrupt the original reasoning architecture. Prompt-based approaches such as CCoT \cite{mitra2024compositional}, KM-COT \cite{mondal2024kam}, and BDoG \cite{zheng2024picture} utilize scene graphs as external knowledge sources without significantly increasing the model's inherent interaction reasoning capabilities. Other methods \cite{liu2023visual,wang2023self} enhance training with region-localized descriptions but fail to effectively capture object interactions. In contrast, our approach integrates fine-grained scene graph information directly into supervised fine-tuning, ensuring models maintain structured knowledge while significantly improving their understanding of object interactions without requiring architectural changes or compromising efficiency.

\section{Interaction-augmented Scene Graph Reasoning}

In this section, we propose \textbf{ISGR (Interaction-augmented Scene Graph Reasoning)}, a framework that enhances vision-language models' ability to perform nuanced interaction reasoning through structured scene graphs. 
As illustrated in Figure \ref{model}, unlike conventional object-centric methods that primarily focus on spatial relationships, ISGR captures functional interactions between objects while maintaining spatial grounding, enabling more coherent and relationally rich scene understanding.

The ISGR can be viewed as an iterative process, where the scene graph is refined iteration by iteration. The output answer at each iteration can be formulated as
\begin{equation}
    \mathcal{T}^{i}=(\mathcal{G}^i,\mathcal{S}, \mathcal{M},\mathcal{F}) 
\end{equation}
where, given a multimodal input $\mathcal{S}=\{Q,I\}$ for a specific question $Q$ and image $I$, the current scene graph $\mathcal{G}^i$ is updated by the multimodal LLM - $\mathcal{M}$ with a set of operation functions $\mathcal{F}$.

It should be noted that ISGR reinforces a long-term memory by tuning $\mathcal{M}$ with the instruction data that comprise the interactional scene graph and queries. As a result, a reasoning answer will be derived directly from $\mathcal{M}$. More details about the memory reinforcement are in section 3.4.

\subsection{Scene Graph Initialization and Abstraction}

In our framework, we first initialize a scene graph that captures both spatial and abstract representations of visual content. This process can be formulated as:
\begin{equation}
\mathcal{G}^0 = f_{\text{init}}(I, Q)
\end{equation}
where $f_{\text{init}}$ represents the initialization function that generates an initial scene graph $\mathcal{G}^0$ from image $I$ with respect to question $Q$.

\subsubsection{Spatial Graph Construction}

The spatial graph construction focuses on identifying objects and their spatial relationships in the scene:
\begin{equation}
\mathcal{G}^s = \langle \mathcal{V}^s, \mathcal{E}^s \rangle
\end{equation}
where $\mathcal{V}^s$ represents entities like ``Building'', ``Trees'', ``Frisbee'', and ``Grass'' as shown in the diagram. Spatial relationships in $\mathcal{E}^s$ include ``behind'', ``on'', ``near'', and ``top of''.

We implement this through prompted inference with the multimodal LLM:
\vspace{10pt}
\begin{mdframed}[hidealllines=true,backgroundcolor=gray!20]
    Generate a spatial scene graph identifying objects and their spatial relationships for: \{\textit{image}\}
\end{mdframed}

\subsubsection{Abstract Graph Construction}

Building upon the spatial graph, we construct an abstract graph that focuses on the contextually relevant elements while reducing noise:
\begin{equation}
\mathcal{G}^a = f_{\text{abstract}}(\mathcal{G}^s, I, Q)
\end{equation}

This abstraction process is guided by three key constraints:
\begin{itemize}
    \item \textbf{Focus Constraint}: Emphasize salient objects that are core to understanding the scene.
    \item \textbf{Relevance Constraint}: Extract only elements that are directly related to the core scene.
    \item \textbf{Disambiguation Constraint}: Resolve ambiguities in object references and relationships within the core scene.
\end{itemize}

This process can be implemented as:
\vspace{10pt}
\begin{mdframed}[hidealllines=true,backgroundcolor=gray!20]
    Create an abstract graph of \{\textit{spatial scene graph}\} that focuses only on elements relevant to the core scene: \{\textit{image}\}. Ensure clarity and disambiguation of entities.
\end{mdframed}

\subsection{Interactional Chain-of-Thoughts (ICoT) Approach}

After constructing the abstract graph, we enhance it with interactional relationships using our proposed Interactional Chain-of-Thoughts (ICoT) approach.

\subsubsection{Interaction Identification and Modeling}

The ICoT process transforms the abstract graph into an interaction graph:
\begin{equation}
\mathcal{G}^t = f_{\text{ICoT}}(\mathcal{G}^a, I, Q)
\end{equation}

As shown in the diagram, this process identifies dynamic relationships such as ``looking at'', ``jumps to'', ``reaches for'', and ``collides'' between entities like ``Player in black'', ``Player in white'', ``Player in red hat'', and ``Frisbee''.

The interaction identification follows this reasoning chain:

\begin{itemize}

    \item \textbf{Subject Identification}: Identify potential actors (e.g., players in different colored clothing)
    \item \textbf{Action Recognition}: Determine actions being performed (e.g., jumping, reaching)
    \item \textbf{Object Identification}: Identify recipients of actions (e.g., the frisbee)
    \item \textbf{Relation Formalization}: Formalize relationships as directional triplets
\end{itemize}

This is implemented through:

\vspace{10pt}
\begin{mdframed}[hidealllines=true,backgroundcolor=gray!20]
Using the \{\textit{abstract graph}\}, identify all interactions between entities that are relevant to the core scene: \{\textit{image}\}. For each interaction, specify the subject, action, and object.
\end{mdframed}

\subsubsection{Further Abstraction with Interaction Knowledge}
The interaction knowledge is used to further abstract the scene, focusing on the most relevant interactions for answering the question:
\begin{equation}
    \mathcal{G}^\text{final} = f_\text{abstract}(\mathcal{G}^t,I,Q)
\end{equation}

This final abstraction is guided by additional constraints:
\begin{itemize}
    \item \textbf{Saliency Constraint}: Emphasize the most important interactions
    \item \textbf{Grounding Constraint}: Ensure interactions are visually grounded in the image
    \item \textbf{Consistency Constraint}: Maintain logical consistency across all represented interactions
\end{itemize}

An exemplar implementation is as follows:

\vspace{10pt}
\begin{mdframed}[hidealllines=true,backgroundcolor=gray!20]
Using the \{\textit{interaction knowledge}\}, further abstract the scene by identifying the most relevant interactions for the core scene: \{\textit{image}\}. For each interaction, specify the subject, action, and object while ensuring adherence to the saliency, grounding, and consistency constraints.
\end{mdframed}

\subsubsection{Querying the Interaction-augmented Graph}

To construct the instruction-tuning dataset with scene-interaction data, we generate the corresponding queries to our interaction-augmented graphs. For example, in the diagram's case, the question ``Who will catch the frisbee?'' requires analyzing interactions between players and the frisbee to determine that ``The people in black will catch the frisbee''.

Our framework supports four types of queries over the interaction-augmented scene graph:

\begin{itemize}
    \item \textbf{Object-Object Queries}: Identify relationships between specific objects
    \begin{equation}
    Q_{o-o}(o_1, o_2) \rightarrow \{r | (o_1, r, o_2) \in \mathcal{E}^t\}
    \end{equation}
    \item \textbf{Subject-Relation Queries}: Find objects related to a subject via a specific relation
    \begin{equation}
    Q_{s-r}(s, r) \rightarrow \{o | (s, r, o) \in \mathcal{E}^t\}
    \end{equation}
    \item \textbf{Relation-Object Queries}: Find subjects that relate to a specific object
    \begin{equation}
    Q_{r-o}(r, o) \rightarrow \{s | (s, r, o) \in \mathcal{E}^t\}
    \end{equation}
    \item \textbf{Comprehensive Queries}: Identify all relationships associated with a specific object.
    \begin{equation}
    Q_{\text{comp}}(o) \rightarrow \{(s, r) | (s, r, o) \in \mathcal{E}^t\}
    \end{equation}
\end{itemize}

This process is carried out by employing:

\vspace{10pt}
\begin{mdframed}[hidealllines=true,backgroundcolor=gray!20]
Using the \{\textit{interaction-augmented graph}\}, generate queries to identify relationships relevant to the core scene: \{\textit{image}\}. For each query type, specify the relevant entities and their interactions while ensuring clarity and contextual relevance.
\end{mdframed}

\subsection{Long-term Memory Reinforcement (LTMR)}

To develop a robust long-term memory for interaction reasoning, we integrate our graph-based approach with memory reinforcement techniques:
\begin{equation}
\mathcal{M}_{\text{enhance}} = f_{\text{memory}}(\mathcal{M}_{\text{base}}, \mathcal{D}_{\text{base}}, \mathcal{D}_{\text{interact}})
\end{equation}
where $\mathcal{M}_{\text{enhance}}$ is the enhanced model, $\mathcal{M}_{\text{base}}$ is the base model, $\mathcal{D}_{\text{base}}$ is the base dataset, and $\mathcal{D}_{\text{interact}}$ is the scene-interaction dataset.

\subsubsection{Dataset Construction and Integration}

We construct our scene-interaction dataset by combining:

\begin{itemize}
    \item Spatial relationships from existing scene graph datasets
    \item Interactional relationships derived through our ICoT approach
    \item Manually verified interaction triplets for quality assurance
\end{itemize}

The integration follows:
\begin{equation}
\mathcal{D}_{\text{interact}} = \{(I_i, Q_i, \mathcal{G}^\text{final}_i\})\}_{i=1}^N
\end{equation}
where $I$ is the input image, $Q$ indicates the generated interactive query from section 3.2.3 and $\mathcal{G}^\text{final}$ represents the final graph derived from our ICoT aproach.

\begin{table*}
  \centering
  \begin{tabular}{lccccccccc}
    \toprule
       Model & LLM & Data Size & VQAv2 & GQA & VizWiz & TextVQA & VSR & MME \\
    \midrule
    EMU & LLaMA-13B & 3.4B & 62 & 46 & 38.3 & - & - & - \\
    OpenFlamingo & MPT-7B & 2B & 52.7 & - & 27.5 & 33.6 & - & - \\
    Qwen-VL & Qwen-7B & 1.5B & 78.2 & 59.3 & 35.2 & 63.8 & - & - \\
    IDEFICS & LLaMA-7B & 354M & 50.9 & - & 35.5 & 25.9 & - & - \\
    InstructBLIP & Vicuna-7B & 130M & - & 49.2 & 34.5 & 50.1 & 54.3 & - \\
    InstructBLIP & Vicuna-13B & 130M & - & 49.5 & 33.4 & 50.7 & 52.1 & 1212.8 \\
    BLIP-2 & Vicuna-13B & 129M & - & 41 & 19.6 & 42.5 & 50.9 & 1293.8 \\
    Shikra & Vicuna-13B & 6.1M & 77.4 & - & - & 25.9 & - & - \\
    MiniGPT-4 & Vicuna-7B & 5M & 32.2 & - & - & - & - & 581.7 \\
    MoE-LLaVA & StableLM-1.6B x4 & 2.2M & 76.7 & 60.3 & 36.2 & 50.1 & - & - \\
    MoE-LLaVA & Phi2-2.7B x4 & 2.2M & 77.6 & 61.4 & 43.9 & 51.4 & - & - \\
    LLaVA v1.5 & Vicuna-7B & 0.6M & 78.5 & 62 & 45.9 & 58.2 & 54.1 & 1352.5 \\
    LLaVA v1.5 & Vicuna-7B & 1.2M & 79.2 & 63.3 & 49.6 & 58.5 & 54.5 & 1256.3 \\
    \midrule
    LLaVA-IRR & Vicuna-7B & 1.2M+500 & 79.6 & 62.6 & 50.8 & 58.9 & 55.6 & 1344.1 \\
    ISGR(SFT)-S(\textbf{Ours}) & Vicuna-7B & 0.8M & 79.4 & 63.4 & 49.5 & 57.4 & 55.7 & \underline{\textbf{1460.1}} \\
    ISGR(SFT)-M(\textbf{Ours}) & Vicuna-7B & 1.3M & \underline{\textbf{80.1}} & \underline{\textbf{63.6}} & 51.3 & 58.4 & \underline{\textbf{61.0}} & 1291.7 \\
    ISGR(SFT+IRR)(\textbf{Ours}) & Vicuna-7B & 1.3M+500 & 79.4 & 62.4 & \underline{\textbf{54.5}} & \underline{\textbf{59.3}} & 60.6 & 1414.1 \\
    \bottomrule
  \end{tabular}
  \caption{Performance comparison across multiple benchmarks. LLM: underlying language model; Data Size: training sample count. We propose three models: ISGR(SFT)-S (0.8M data), ISGR(SFT)-M (1.3M data), and ISGR (SFT+IRR). The various benchmarks (VQAv2, GQA, VizWiz, TextVQA, VSR, and MME) assess different aspects of visual reasoning capabilities across diverse tasks.}
  \label{tab:multimodel}
  \vspace{-10pt}
\end{table*}

\subsubsection{Memory Reinforcement Training}

Our memory reinforcement training involves two phases:

\begin{enumerate}
    \item \textbf{Supervised Fine-tuning (SFT)} using the interaction augmented scene graphs:
    \begin{equation}
    \mathcal{L}_{\text{SFT}} = \mathbb{E}_{(I,Q,\mathcal{G}) \sim \mathcal{D}} \left[ -\log P_{\mathcal{M}}(\mathcal{G} | I, Q) \right]
    \end{equation}
    \item \textbf{Interaction Reasoning Reinforcement (IRR)} through a reward-based mechanism:
    \begin{equation}
    \mathcal{L}_{\text{IRR}} = \mathbb{E}_{(I,Q,A) \sim \mathcal{D}} \left[ R(A_{\text{pred}}, A_{\text{gt}}) \right]
    \end{equation}
\end{enumerate}

The reward function $R$ evaluates both the quality of the interaction graph and the correctness of the final answer. For each image-question pair, the model generates $K$ candidate responses $\{y_1, y_2, \ldots, y_K\}$, each evaluated using a specialized reward function targeting relational accuracy:
\begin{equation}
R(y_k) = \lambda_1 \cdot \mathcal{F}_{\text{focus}}(y_k) + \lambda_2 \cdot \mathcal{F}_{\text{disamb}}(y_k) - \lambda_3 \cdot \mathcal{F}_{\text{rele}}(y_k)
\end{equation}

This function comprises three key components:
\begin{itemize}
    \item $\mathcal{F}_{\text{focus}}(y_k)$: Evaluates how well the response focuses on central entities relevant to the question
    \item $\mathcal{F}_{\text{disamb}}(y_k)$: Measures the clarity and lack of ambiguity in entity references
    \item $\mathcal{F}_{\text{rele}}(y_k)$: Penalizes irrelevant information that may distract from the core reasoning task
\end{itemize}

Through extensive experimentation, we determined that the optimal hyperparameter values are $\lambda_1 = 0.4$, $\lambda_2 = 0.4$, and $\lambda_3 = 0.2$, effectively balancing the competing constraints of focus, disambiguation, and relevance in the generated scene graphs.

\section{Experiments}

Our experimental evaluation is designed to systematically analyze how our proposed approach addresses the key limitations of traditional scene graph construction methods outlined in the introduction. 
Specifically, we assess: (1) the effectiveness of our interaction-augmented scene graphs compared to conventional spatial-only graphs; 
(2) the benefits of our summarize-and-align approach for reducing contextual drift; and 
(3) the impact of long-term memory reinforcement via GRPO on generalizing interactional reasoning to unseen data.

\subsection{Experimental Setup}

\noindent \textbf{Dataset Construction.}
To support interaction-focused scene graph learning, we constructed a specialized dataset combining multiple sources: LLaVA-v1.5-mixed-665k\cite{liu2023visual}, 176K images from OpenImages\cite{kuznetsova2020open} with manually annotated scene graphs, LVIS-Instruct-4V\cite{wang2023see}, and LRV-Instruct\cite{liu2023mitigating}. 
We created two variants for SFT: an 841K dataset combining LLaVA-v1.5-mixed-665k and OpenImages, and a larger 1,371K dataset that incorporates LVIS-Instruct-4V and LRV-Instruct. Both variants include 300K interaction-augmented scene graph data to ensure robust interaction reasoning capabilities. Additionally, we utilized a separate set of 500 high-quality interaction instructions specifically designed for interaction reasoning reinforcement.

\noindent \textbf{Implementation Details.} We trained our models on 8 NVIDIA A100 GPUs (40GB) using LLaVA-v1.5 (7B) architectures. 
Supervised fine-tuning (SFT) was performed from pre-trained checkpoints following official protocols, with the per-device batch size reduced from 16 to 8 due to hardware constraints. 
Our models underwent fine-tuning on the interaction-augmented dataset, followed by interaction reasoning reinforcement using 500 high-quality interaction instruction examples to enhance interaction reasoning capabilities.

\subsection{Evaluation Framework}

We carefully selected a diverse suite of benchmarks to comprehensively evaluate both general vision-language capabilities and specific interactional reasoning skills:
\begin{itemize}
    \item \textbf{General VL Understanding}: VQAv2\cite{goyal2017making} (diverse question types), VizWiz\cite{gurari2018vizwiz} (real-world accessibility questions), and TextVQA\cite{singh2019towards} (text-focused reasoning)
    
    \item \textbf{Spatial \& Relational Understanding}: GQA\cite{hudson2019gqa} (compositional spatial reasoning), VSR\cite{liu2023vsr} (visual spatial reasoning)
    
    \item \textbf{Real-world Interaction Understanding}: RealWorldQA\cite{grok15v} (practical spatial understanding), MMT-Bench\cite{ying2024mmt} (recognition, localization, and reasoning)
    
    \item \textbf{Compositional Reasoning}: SEEDBench\cite{li2024seed} (interaction, spatial and temporal understanding), A-Bench\cite{zhang2024bench} (scene understanding in synthetic images)
\end{itemize}

\subsection{Baseline Models for Comparison}
To comprehensively evaluate the performance of our model, we compare it against a diverse set of strong multi-modal baselines across different model scales and dataset sizes:

We compare against a diverse set of vision-language models spanning different scales. Large-capacity models such as EMU 2\cite{sun2023emu}, OpenFlamingo\cite{awadalla2023openflamingo}, Shikra\cite{chen2023shikra}, BLIP-2\cite{li2023blip} and InstrutBlip\cite{dai2023instructblip} leverage extensive pretraining for strong generalization. 
We also include mid-sized models like IDEFICS\cite{hua2024talk}, MiniGPT-4\cite{zhu2023minigpt}, Qwen-VL\cite{Qwen2-VL}, MoE-LLaVA\cite{lin2024moe} and LLaVA v1.5\cite{liu2023visual}, which offer competitive performance under moderate resource settings.

Our proposed models include: 

\begin{itemize}
    \item \textbf{LLaVA-IRR}: Built on the baseline LLaVA-v1.5 (Vicuna-7B) model and directly fine-tuned using our Interaction Reasoning Reinforcement, bypassing the interaction-augmented scene graph training stage.
    \item \textbf{ISGR(SFT)}: Built on the Vicuna-7B architecture and fine-tuned with our interaction-augmented scene graph dataset through supervised fine-tuning. We provide two versions: (1) ISGR(SFT)-S trained with 0.8M augmented data, and (2) ISGR(SFT)-M trained with a larger 1.3M dataset for enhanced performance.
    \item \textbf{ISGR(SFT+IRR)}: An enhanced version of ISGR(SFT)-M that undergoes further optimization through Interaction Reasoning Reinforcement to strengthen scene interaction reasoning capabilities. 
\end{itemize}

\begin{figure}[t]
    \centering
    \includegraphics[width=\columnwidth]{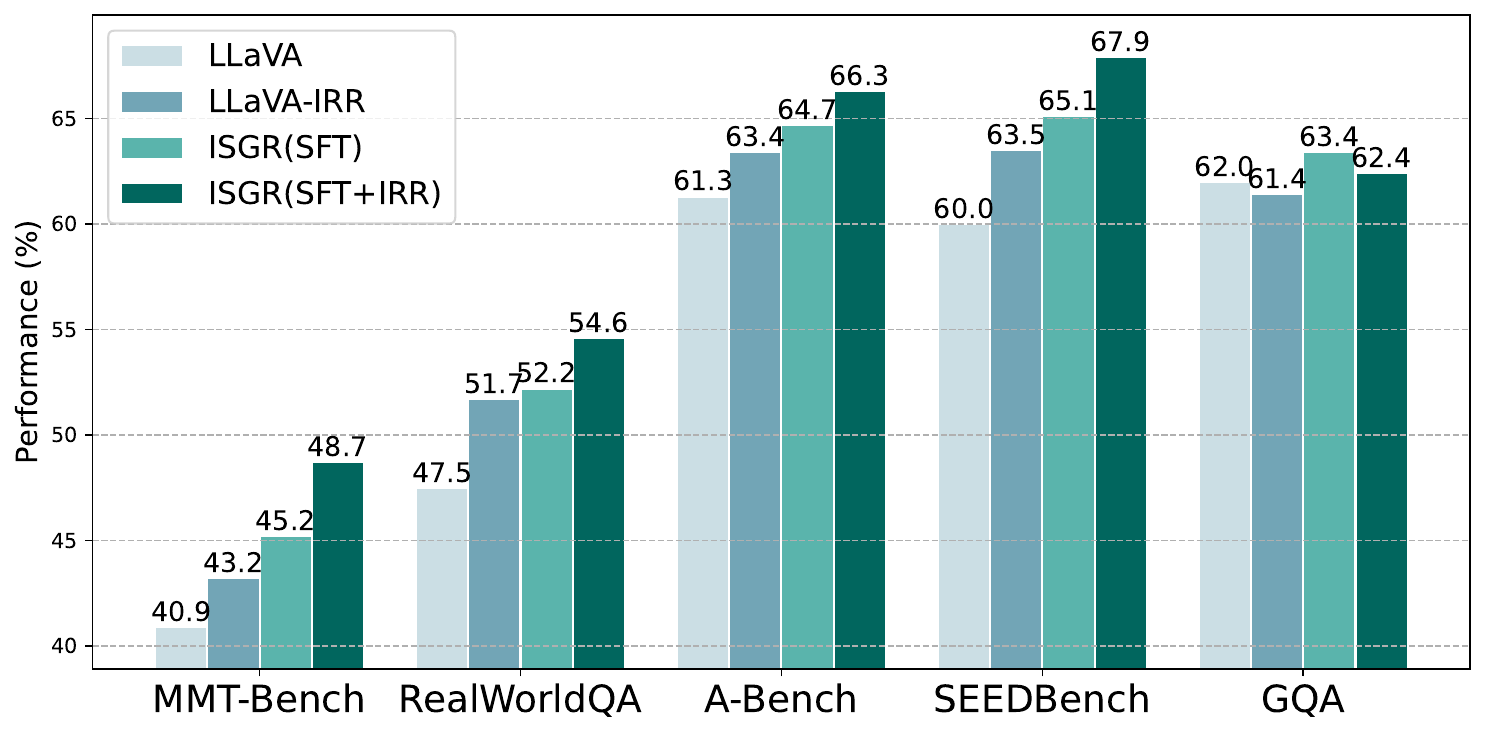}
    \caption{Performance comparison on scene reasoning benchmarks. Our proposed models (ISGR(SFT) and ISGR(SFT+IRR)) consistently outperform baseline models (LLaVA and LLaVA-IRR) across diverse benchmarks measuring different aspects of scene understanding.}
    \vspace{-0.1cm}
    \label{fig:scene_reasoning}
    \vspace{-10pt}
\end{figure}

\subsection{Main Results}
\textbf{Multimodal Question Answering}. 
Table~\ref{tab:multimodel} presents the overall performance of our models on standard multi-modal understanding benchmarks, with our proposed model ISGR achieving even better performance, reaching state-of-the-art results.
Notably, ISGR(SFT)-S trained with only 0.8M interaction-augmented data enables more efficient learning from fewer examples by providing richer supervisory signals, which manages to remain competitive with the LLaVA-v1.5 baseline (1.2M data). 
Further scaling to 1.3M scene-graph enriched samples allows ISGR(SFT)-M to achieve even better performance on
Moreover, while our ISGR(SFT+IRR) model is specifically designed for interaction reasoning rather than text-centric tasks, it still demonstrates impressive performance on text-intensive benchmarks such as TextVQA (+1.1\%). 

Moreover, the model excels in scene understanding datasets, achieving outstanding performance on VizWiz (+5.0\%), and VSR (+7.4\%), demonstrating its strong generalization capabilities across different types of visual reasoning scenarios.

\noindent \textbf{Scene Reasoning.} 
To thoroughly evaluate our model's capabilities on complex scene understanding tasks, we conducted extensive testing across multiple specialized benchmarks that assess different aspects of visual reasoning. 
Table~\ref{fig:scene_reasoning} presents these results, showing substantial improvements across all scene reasoning benchmarks.

Our ISGR(SFT) model demonstrates consistent gains over the LLaVA-v1.5 baseline, with the most notable improvements on SEEDBench (+5.1\%), RealWorldQA (+4.7\%), and MMT-Bench (+4.3\%). 
These improvements directly validate the effectiveness of our interaction augmented scene graph approach. 
The strong performance on GQA (+1.4\%) confirms enhanced spatial reasoning capabilities, while the gains on A-Bench (+3.4\%) demonstrate better generalization to novel and synthetic scenes. 

ISGR(SFT+IRR), with its interaction reasoning reinforcement, pushes performance boundaries even further. 
The most substantial improvements are observed in benchmarks requiring nuanced interaction reasoning: SEEDBench (+7.9\%), MMT-Bench (+7.8\%), and RealWorldQA (+7.1\%). 

\begin{figure}[t]
    \centering
    \includegraphics[width=\columnwidth]{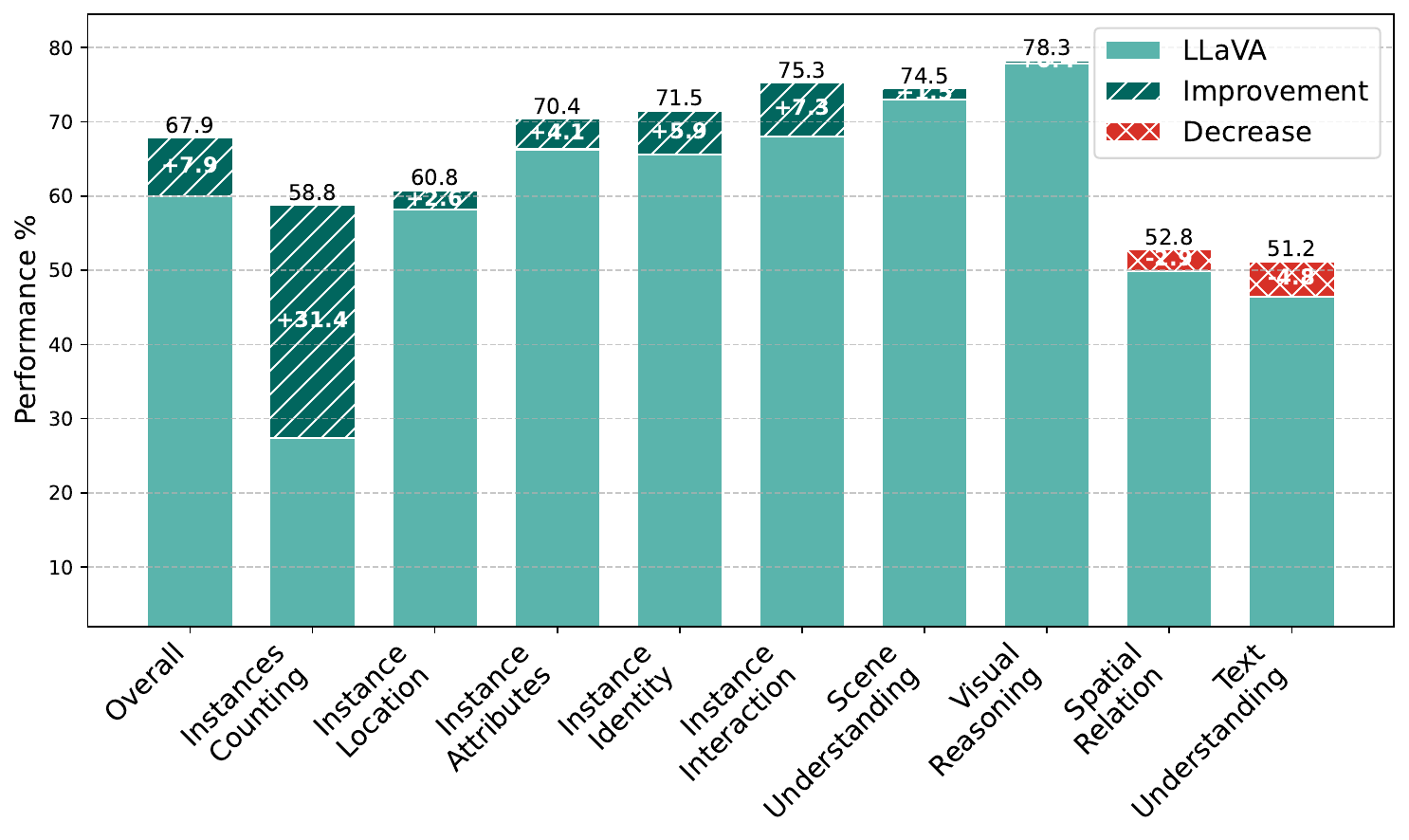}
    \caption{Category Performance Comparison on SEEDBench. ISGR(SFT+IRR) shows significant improvements over the LLaVA-v1.5 baseline across most categories.}
    \label{fig:seedbench_class}
\end{figure}

\subsection{Ablation Study}
\noindent \textbf{Overcoming Limitations of Spatial-Only Relationships.} One of the key limitations identified in the introduction is the overreliance of existing methods on spatial relationships. 
Figure~\ref{fig:seedbench_class} provides a detailed analysis of how our approach enhances different aspects of visual understanding on the SEEDBench dataset. 
The most substantial improvement is observed in Instance Interaction (+7.3\%), directly validating our approach's effectiveness in capturing dynamic relationships between objects. 

Additionally, our approach demonstrates remarkable improvement in Instance Counting (+31.4\%), suggesting that modeling interactions helps the model better distinguish and enumerate individual instances in the scene. 
The gains in Instance Identity (+5.9\%) and Instance Attributes (+4.1\%) further indicate that understanding interactions helps the model form more comprehensive object representations. 
Collectively, these improvements contribute to a significant overall performance gain (+7.9\%) across all categories.

Interestingly, we observe slight decreases in Spatial Relation and Text Understanding categories. 
This trade-off suggests that while our model excels at interaction-focused reasoning, extremely fine-grained spatial relationship modeling may be marginally affected as the model prioritizes functional over purely positional relationships.

\begin{table}[!t]
  \centering
  \resizebox{\linewidth}{!}{
    \begin{tabular}{ccccccccccc}
    \toprule
    \textbf{IT} & \textbf{\( Q_{oo} \)} & \textbf{\( Q_{sro} \)} & \textbf{\( Q_{cs} \)} & \textbf{VQA$^{v2}$} & \textbf{GQA} & \textbf{RWQA} & \textbf{MMT} & \textbf{A-Bench} & \textbf{Avg.} \\
    \midrule
    \multirow{4}[2]{*}{0.8M} & \ding{55} & \ding{55} & \ding{55} & 78.5  & 62  & 47.5  & 40.9  & 61.3  & 58.04 \\
           & \ding{51} & \ding{55} & \ding{55} & 79.0  & 63.1  & 36.1  & 35.4  & 59.9  & 54.70 \\
           & \ding{55} & \ding{51} & \ding{51} & 79.3  & 63.2  & 52.8  & 43.2  & 63.1  & 60.32 \\
           & \ding{51} & \ding{51} & \ding{55} & 79.3  & 63.2  & 54.5  & 43.7  & 64.6  & 61.06 \\
    \rowcolor[HTML]{E6EFDB}
           & \ding{51} & \ding{51} & \ding{51} & 79.4  & 63.4  & 52.2  & 45.2  & 64.7  & 61.54 \\
    \midrule
    \multirow{4}[2]{*}{1.3M} & \ding{55} & \ding{55} & \ding{55} & 80.0  & 63.3  & 43.0  & 30.3  & 45.0  & 52.34 \\
           & \ding{51} & \ding{55} & \ding{55} & 80.2  & 63.1  & 45.4  & 34.4  & 59.3  & 56.48 \\
           & \ding{55} & \ding{51} & \ding{51} & 80.1  & 63.1  & 45.5  & 43.6  &  62.1 & 58.9 \\
           & \ding{51} & \ding{51} & \ding{55} & 80.2  & 63.5  & 47.9  & 44.8  & 63.3  & 59.94 \\
    \rowcolor[HTML]{E6EFDB}
           & \ding{51} & \ding{51} & \ding{51} & 80.1  & 63.6  & 53.2  & 48.1  & 64.8  & 61.96 \\
    \bottomrule
    \end{tabular}%
  }
    \caption{Ablation study on instruction categories for Summarize-and-Align graph within the ISGR (SFT) model. Highlighted rows (green) demonstrate that incorporating all query types yields the best overall performance, confirming the complementary nature of different interaction-focused instruction categories.}
    \label{tab:ablation_Q}
    \vspace{-15pt}
\end{table}%

\begin{table}[t]
  \centering
  \resizebox{\linewidth}{!}{
    \begin{tabular}{cccccccccc}
    \toprule
    \textbf{IT} & \textbf{Rel.} & \textbf{Dis.} & \textbf{Foc.} & \textbf{GQA} & \textbf{RWQA} & \textbf{MMT} & \textbf{A-Bench} & \textbf{SEEDB} & \textbf{Avg.} \\
    \midrule
    \multirow{4}[2]{*}{1.3M} & \ding{51} & \ding{55} & \ding{55} & 62.3  & 53.5  & 48.6  & 65.4 & 67.2 & 59.40 \\
           & \ding{55} & \ding{51} & \ding{55} & 62.2 & 53.8 & 49.1 & 65.2 & 67.1 & 59.48 \\
           & \ding{55} & \ding{55} & \ding{51} & 62.2 & 53.4 & 48.6 & 65.4 & 66.7 & 59.26 \\
    \rowcolor[HTML]{E6EFDB}
           & \ding{51} & \ding{51} & \ding{51} & 62.4 & 54.6 & 48.7 & 66.3 & 67.9 & 59.98 \\
    \bottomrule
    \end{tabular}%
  }
  \caption{Ablation study on Long-Term Memory Reinforcement (LTMR) showing the impact of different reward components (Relevance, Disambiguation, Focus) on ISGR (SFT+IRR) model performance across various benchmarks.}
  \vspace{-0.7cm}
  \label{tab:ablation_Grpo}
\end{table}%

\noindent \textbf{Summarize-and-Align Graph Construction Effectiveness.}
To quantitatively evaluate the effectiveness of our summarize-and-align approach, we conducted a comprehensive ablation study within the ISGR (SFT) model, examining the contribution of different query types in our Scene-Interaction dataset. 
We conducted ablation studies on three instruction categories in our dataset: Object-Object queries (\(Q_{oo}\)), Subject-Relation-Object queries (\(Q_{sro}\)), and Comprehensive Subject queries (\(Q_{cs}\)), which together capture different dimensions of scene relationship reasoning.

Table~\ref{tab:ablation_Q} presents the results of our ablation study across two data sizes (0.8M and 1.3M). 
The consistent pattern across all benchmarks clearly demonstrates that our complete approach—incorporating all three query types—significantly outperforms partial implementations. 
With the full 1.3M dataset, using all components achieves the highest average performance (61.96\%) across the five benchmarks, compared to just 52.34\% when using none of these specialized queries.
Particularly noteworthy is the performance on scene-specific reasoning benchmarks (RealWordQA, MMT-Bench, and A-Bench), where the gains are most substantial. 
For instance, on MMT-Bench, the full approach achieves 48.1\% compared to 30.3\% for the baseline—a remarkable 17.8\% point improvement. 

\begin{figure*}
    \centering
    \includegraphics[width=7in]{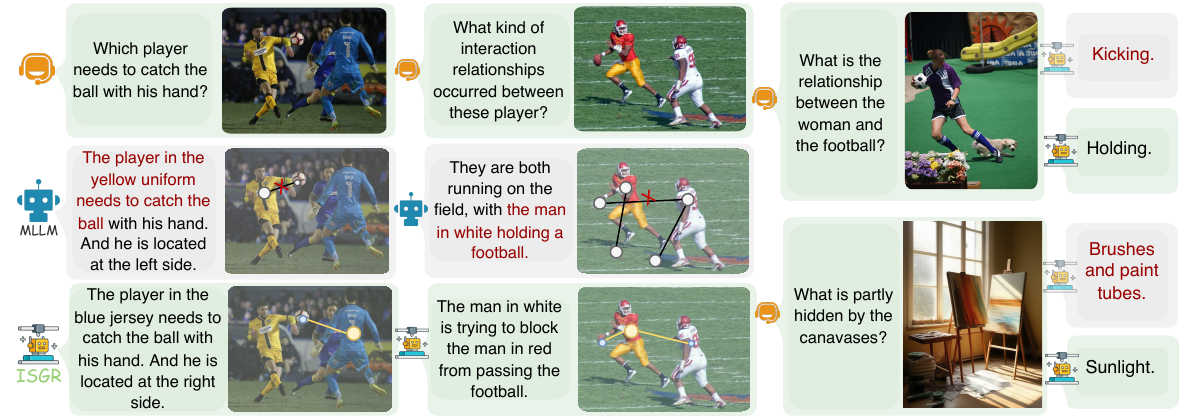}
    \caption{A case study for our proposed ISGR framework: (Left) Limitations of spatial reasoning where models provide contradictory answers based solely on proximity; (Medium) Scene graph focusing enables accurate identification of functional interactions; (Right) Long-term memory reinforcement enhances subtle relationship identification and generalization to novel scenes, including unseen relationship types in generated images.}
    \vspace{-0.1cm}
    \label{fig:case}
\end{figure*}

Remarkably, our experiments reveal the complementary nature of the different query types. 
Using \( Q_{oo} \) alone with the 0.8M dataset actually degrades performance on scene reasoning tasks compared to the baseline (-3.34\%), suggesting that object-object relationships in isolation may lead to focus drift without the constraining context provided by the other query types. 
However, when combined with \( Q_{sro} \) and \( Q_{cs} \), performance improves dramatically, confirming that comprehensive relational modeling is necessary for effective summarize-and-align graph construction.

\noindent \textbf{Long-Term Memory Reinforcement.}
long-term memory reinforcement(LTMR) is designed to enhance the model's ability to retain and generalize interaction patterns across diverse visual scenes. 
Figure~\ref{fig:scene_reasoning} presents our comprehensive comparison of interaction-augmented tuning and LTMR impacts across multiple benchmarks. 
While fine-tuning with our Graph-Interaction dataset already provides substantial improvements over the baseline (+1.4\% on GQA, +4.7\% on RealWorldQA, +4.3\% on MMT-Bench), incorporating LTMR further enhances performance significantly on benchmarks requiring complex interaction reasoning: RealWorldQA (+7.1\%), MMT-Bench (+7.8\%), and SEEDBench (+7.9\%).

Moreover, the effectiveness of LTMR is particularly evident on benchmarks that test real-world spatial understanding and fine-grained interaction reasoning. 
While we observe a slight performance decrease on GQA (-1.0\%) when adding LTMR to the fine-tuned model, the substantial gains on more challenging benchmarks demonstrate LTMR's ability to enhance generalization to complex interaction patterns. 
This trade-off suggests that LTMR optimization slightly shifts the model's focus from purely spatial relationships toward more functional and causal interactions, which aligns with our goal of improving interaction-based reasoning.

To evaluate the effectiveness of LTMR and understand the contribution of different reward components, we conducted a detailed ablation study presented in Table~\ref{tab:ablation_Grpo}. 
We examined three key reward components: Relevance (Rel.), Disambiguation (Dis.), and Focus (Foc.) through our ablation experiments.

Our results demonstrate that all three reward components contribute to the model's reasoning capabilities, with the full combination yielding the best average performance (59.98\%) across all benchmarks. 
Notably, the Disambiguation component shows the strongest individual effect (59.48\%), highlighting the critical importance of clearly identifying subject-object roles in interaction reasoning—a core limitation we identified in conventional approaches.

\subsection{Case Study} 

Our interaction-augmented scene graph approach demonstrates significant advantages over traditional spatial-only methods by effectively capturing functional relationships within scenes. 
This is clearly illustrated in Figure~\ref{fig:case}, which showcases our multi-level reasoning framework.

The left case draws a soccer scene from the SEEDBench benchmark, where a conventional model, relying solely on spatial proximity, incorrectly identifies the yellow player as the one who needs to catch the ball. 
In contrast, our model employs contextual understanding by incorporating soccer-specific rules, correctly identifying the goalkeeper as the only player allowed to handle the ball.

The example on the middle comes from the MMT-Bench benchmark require focus-guided reasoning. 
The baseline model produces an unfocused response, stating, "They are both running on the field, with the man in white holding a football," which reveals both attention drift and factual inaccuracies. 
Our approach, however, establishes critical interaction points and applies selective attention filtering, leading to the correct identification: "The man in white is trying to block the man in red from passing the football."

The right case demonstrates our Long-Term Memory Reinforcement (LTMR) mechanism's impact. 
While the ISGR(SFT) model struggles to generalize interaction relationships (like distinguishing between "kicking" and "holding" a ball), our ISGR(SFT+IRR) model activates relevant interaction patterns from previously processed examples for accurate identification. 
The example from A-Bench tests generalization in a generated image scenario, showing our approach handles both different data domains and unseen interaction relationships effectively.

\section{Conclusion}

We presented Interaction-augmented Scene Graph Reasoning (ISGR), a framework that enhances vision-language models' ability to reason about complex interactions in visual scenes. 
By extending beyond traditional spatial-only representations to capture functional relationships, our approach effectively addresses focus drift and contextual ambiguity issues. 
Experiments demonstrate that our models achieve strong performance across diverse benchmarks with less training data, while our long-term memory reinforcement mechanism further improves generalization to novel interaction scenarios. 
These results confirm the value of structured relational modeling in visual reasoning tasks.



\bibliographystyle{ACM-Reference-Format}
\bibliography{sample-base}

\appendix

\section{More Experimental Details}

\algnewcommand{\LineComment}[1]{\State \(\triangleright\) #1}

\subsection{Algorithm for ISGR}

For a better understanding of ISGR, an algorithmic procedure has been formulated in Algorithm 1. 

\begin{algorithm}[htb]
    \caption{ISGR (Interaction-augmented Scene Graph Reasoning)}
    \begin{algorithmic}
    \Require{Input $S$ = (question $Q$, Wild Image $I_{max}$), Multimodal LLM $\mathcal{M}$.}

    \For{$I = 1$ to $I_{max}$}
    \noindent \textbf{Initialize} scene graph $\mathcal{G}^0 \gets f_{\text{init}}(I, Q)$.
     
    \LineComment{Construct Spatial Graph}
    \State $\mathcal{G}^s \gets f_{\text{spatial}}(I, Q, \mathcal{M})$
    
    \LineComment{Update Abstract Graph}
    \State $\mathcal{G}^a \gets f_{\text{abstract}}(\mathcal{G}^s, I, Q)$
    
    \LineComment{Generate Interaction Graph}
    \State $\mathcal{G}^t \gets f_{\text{ICoT}}(\mathcal{G}^a, I, Q)$
    
    \LineComment{Further Abstraction}
    \State $\mathcal{G}^\text{final} \gets f_{\text{abstract}}(\mathcal{G}^t, I, Q)$
    
    \EndFor
    
    \LineComment{Supervised Fine-Tuning(SFT)}
    \State $\mathcal{L}_{\text{SFT}} \gets \mathbb{E}_{(I,Q,\mathcal{G}^\text{final})} \left[ -\log P_{\mathcal{M}}(\mathcal{G}^\text{final} | I, Q) \right]$

    \LineComment{Interaction Reasoning Reinforcement(IRR)}
    \State $\mathcal{L}_{\text{IRR}} \gets \mathbb{E}_{(I,Q,A) \sim \mathcal{D}} \left[ R(A_{\text{pred}}, A_{\text{gt}}) \right]$

    \State \textbf{Output} answer based on ISGR and $Q$.

    \end{algorithmic}
\end{algorithm}

\begin{table}[htb]
\small
    \centering
    \begin{tabular}{c |c|c}
    \hline
        Dataset & Source & Size\\
        \hline
        LLaVA-v1.5-mixed-665k \cite{liu2023visual} & LLaVA & 665K  \\
        LVIS-Instruct-4V \cite{wang2023see} & LVIS & 300K  \\
        LRV-Instruct \cite{liu2023mitigating} & LRV & 300K  \\
        OpenImages \cite{kuznetsova2020open}  & OpenImages & 176K \\
        Interaction-Augmented & OpenImages + Exist Set & 300K \\
        \hline
        \textbf{Small-Scale Variant}  & LLaVA + Interaction-Augmented & 841K \\
        \textbf{Medium-Scale Variant} & Whole & 1,371K \\
        \hline
        IRR Instruction & Curated Set & 500 \\
        \hline
    \end{tabular}
    \caption{Statistics of the constructed dataset for interaction-focused scene graph learning. Each variant includes interaction-augmented scene graph data to support robust reasoning capabilities.}
    \label{tab:dataset_construction}
\end{table}

\begin{table}[htbp]
  \centering
  \begin{tabular}{c|c}
    \hline
    Setting & Value \\
    \hline
    Language Model (LLM) & Vicuna-7B \\
    Vision Encoder & CLIP-L/14 \\
    Hardware Requirement & 8x A100 (40GB) \\
    \hline
    Truncation Mode & Left \\
    Number of Beams & 1 \\
    Batch Size & 8 \\
    Temperature & 0.2 \\
    Top-p & 0.9 \\
    Data Type & float16 \\
    \hline
    Image Resolution & 224x224 \\
    Maximum Input Length & 512 \\
    Maximum Output Length & 128 \\
    \hline
    Train Time for ISGR(SFT)-S & ~13 hours \\
    Train Time for ISGR(SFT)-M & ~25 hours \\
    Train Time for ISGR(SFT+IRR) & ~ 34 min \\
    \hline
    Inference Time for VQAv2 & ~7.1 s/sample \\
    Inference Time for GQA & ~8.9 s/sample \\
    Inference Time for SEEDBench & ~9.2 s/sample \\
    Inference Time for MMT-Bench & ~10.5 s/sample \\
    \hline
  \end{tabular}
  \caption{ISGR Model Fine-Tuning and Inference Settings}
  \label{tab:llava7b_config}
\end{table}

\subsection{Statistics of Datasets}

Table~\ref{tab:dataset_construction} summarizes the key statistics of the constructed dataset designed for interaction-focused scene graph learning. This dataset integrates multiple sources, including LLaVA-v1.5-mixed-665k, LVIS-Instruct-4V, LRV-Instruct, and OpenImages, totaling 1,371K instances in the medium-scale variant. The small-scale variant includes 841K instances, combining LLaVA with interaction-augmented data. Additionally, we have a curated set of 500 high-quality instructions specifically designed for Interaction Reasoning Reinforcement (IRR). Each variant of the dataset enhances the model's ability to understand and reason about complex interactions within visual scenes, ensuring robust performance in various visual reasoning tasks.

\subsection{Model Deployment}

The specifics of model deployment and hyperparameter configurations for the ISGR model are detailed in Table \ref{tab:llava7b_config}, highlighting hardware requirements, training parameters, and inference times across various benchmarks.

\subsection{Prompts}
\subsubsection{Spatial initialization}
\begin{mdframed}[hidealllines=true,backgroundcolor=gray!20]
You are an AI assistant. Generate a spatial scene graph identifying objects and their spatial relationships in the given image.

\noindent Use the format of relationship triples: \textbf{<subject, relation, object>}.

\noindent Example Output:
- <person, on, chair>
- <table, next to, chair>

\noindent Input: \textbf{\{image\}}
\noindent Output: \textbf{\{spatial scene graph\}}
\end{mdframed}

\subsubsection{abstract graph}
\begin{mdframed}[hidealllines=true,backgroundcolor=gray!20]
You are an AI assistant. Based on the given \textit{spatial scene graph}, create an abstract version of the graph that focuses only on elements relevant to the core scene described in the image.

\noindent Your task:
\begin{itemize}
    \item Filter out less important or background elements.
    \item Keep only the essential objects and their spatial relationships that define the main activity or layout of the scene.
    \item Ensure all entities are clearly named and unambiguous.
\end{itemize}

\noindent Format:
\begin{itemize}
    \item <subject, relation, object>
    \item Example: <person, sitting on, chair>
\end{itemize}

\noindent Input: \textbf{\{spatial scene graph\}, \{image\}}

\noindent Output: \textbf{\{abstract graph\}}
\end{mdframed}

\subsubsection{Interaction Knowledge}
\begin{mdframed}[hidealllines=true,backgroundcolor=gray!20]
You are an AI assistant. Using the \textit{abstract graph}, identify all interactions between entities that are relevant to the core scene depicted in the image.

\noindent Your task:
\begin{itemize}
    \item Analyze the abstract graph to extract meaningful interactions.
    \item For each interaction, specify the subject, action, and object.
    \item Ensure that all entities are clearly defined and unambiguous.
\end{itemize}

\noindent Format:
\begin{itemize}
    \item <subject, action, object>
    \item Example: <player in blue, passing, football>
\end{itemize}
\noindent Input: \textbf{\{abstract graph\}, \{image\}}

\noindent Output: \textbf{\{interaction knowledge\}}
\end{mdframed}

\subsubsection{Interaction Graph}
\begin{mdframed}[hidealllines=true,backgroundcolor=gray!20]
You are an AI assistant. Using the \textit{interaction knowledge}, further abstract the scene by identifying the most relevant interactions for the core scene depicted in the image.

\noindent Your task:
\begin{itemize}
    \item Focus on the essential interactions that define the dynamics of the scene.
    \item For each interaction, specify the subject, action, and object.
    \item Ensure clarity by adhering to the saliency, grounding, and consistency constraints.
\end{itemize}

\noindent Format:
\begin{itemize}
    \item <subject, action, object>
    \item Example: <goalkeeper, catching, ball>
\end{itemize}
\noindent Input: \textbf{\{interaction knowledge\}, \{image\}}

\noindent Output: \textbf{\{interaction graph\}}
\end{mdframed}

\subsubsection{Querying the Interaction-augmented graph}
\begin{mdframed}[hidealllines=true,backgroundcolor=gray!20]
You are an AI assistant, and you are seeing a single scene graph relationships. The scene graph describes relationships between objects with their bounding box coordinates.

\noindent Given these relationships {\textbf{RELATIONSHIPS TRIPLES}}

\noindent Create \textbf{4 natural QA pairs} about these relationships. You can: 

\noindent 1. \textbf{Q}: What is the relationship between object1[bbox] and object2[bbox]? 

\textbf{A}: object1 relation object2. 

\noindent 2. \textbf{Q}: What does object1[bbox] relation? 

\textbf{A}: object1 relation object2[bbox]. 

\noindent 3. \textbf{Q}: What is relation by object2[bbox]? 

\textbf{A}: object1[bbox] relation object2. 

\noindent 4. \textbf{Q}: What objects have a relationship with object1[bbox]? 

\textbf{A}: object1 relation1 object2[bbox], relation2 object3[bbox], etc. 

\noindent When creating questions:

\noindent \noindent - Focus on the main subject as provided in the scene graph

\noindent - Include bounding box coordinates in the question for specific object identification

\noindent - In the answer, only include bounding box coordinates for objects that weren't specified with coordinates in the question

\noindent - Use the exact relationship and object names as provided in the scene graph

\noindent - Only ask questions that can be definitively answered using the provided scene graph information

\noindent Provide clear and precise answers that directly reflect the relationships shown in the scene graph. Each answer should be specific and correspond exactly to the information available in the scene graph data.

\noindent Input: \textbf{\{interaction graph\}, \{image\}}

\noindent Output: \textbf{\{interaction instruction\}}
\end{mdframed}

\subsection{More Ablation Study}

\noindent \textbf{Impact of Spatial Grounding.}
To investigate the contribution of grounding bounding boxes in scene graph understanding, we conducted ablation experiments by removing bounding box information from both our Long-term Memory Reinforcement(LTMR) training processes.
Table~\ref{tab:ablation_B} presents the comparative results across multiple benchmarks.

The experimental results demonstrate that grounding information consistently improves model performance across all evaluated datasets. 
For the base ISGR(SFT), removing bounding box information leads to performance drops ranging from 0.5\% to 2.2\%. 
This decline is particularly noticeable on specialized visual reasoning benchmarks like MMT-Bench (-1.5\%) and SEEDBench (-2.2\%), suggesting that spatial grounding plays a crucial role in complex visual understanding tasks.

When examining our ISGR(SFT+IRR) model, we observe a similar pattern of performance decline when bounding boxes are removed, with drops of 0.9\% on GQA and 2.0\% on RealWorldQA. 
This highlights the importance of spatial grounding for scene interaction reasoning. 
However, it is notable that ISGR(SFT+IRR) without bounding boxes still outperforms the fully-equipped ISGR(SFT) model on most benchmarks (except GQA), demonstrating that LTMR can elicit strong reasoning capabilities even without explicit spatial grounding.

These findings underscore the significance of spatial grounding in visual reasoning tasks while revealing that our SFT+IRR approach can stimulate powerful reasoning abilities even in its absence. 
Nevertheless, when spatial grounding is provided within the LTMR, it further enhances the model's capabilities in scene interaction reasoning, suggesting that the combination of LTMR and explicit spatial information yields the most robust visual understanding performance.

\begin{table}
  \caption{Impact of spatial grounding on model performance. While spatial grounding consistently improves results, LTMR shows robust reasoning capabilities even without explicit spatial information.}
  \resizebox{\linewidth}{!}{
  \label{tab:ablation_B}
  \centering
  \begin{tabular}{lccccc}
    \toprule
    Model & GQA & RealWorldQA & MMT-Bench & A-Bench & SEEDBench \\
    \midrule
    ISGR(SFT) & 63.6 & 52.2 & 45.2 & 64.7 & 65.1 \\
    -SG/Bbox & 62.9\textcolor{red}{(-0.7)} & 51.7\textcolor{red}{(-0.5)} & 43.7\textcolor{red}{(-1.5)} & 63.1\textcolor{red}{(-1.6)} & 62.9\textcolor{red}{(-2.2)} \\
    \midrule
    ISGR(SFT+IRR) & 62.4 & 54.6 & 48.7 & 66.3 & 67.9 \\
    -SG/Bbox & 61.5\textcolor{red}{(-0.9)} & 52.6\textcolor{red}{(-2.0)} & 47.8\textcolor{red}{(-0.9)} & 65.1\textcolor{red}{(-1.2)} & 66.4\textcolor{red}{(-1.5)} \\
    \bottomrule
  \end{tabular}
  }
\end{table}

\end{document}